\documentclass[10pt, a4paper]{article}

\usepackage{graphicx}
\usepackage{array}
\usepackage{xcolor}
\usepackage{listings}
\usepackage{mdframed}
\usepackage{lipsum} % For dummy text

\usepackage{booktabs}
\usepackage[final]{lrec2026} % this is the new style
\title{Causal Connections: Leveraging Multilingual Fine-Tuning for Financial QA@FinCausal 2026}

\name{Akash Kumar Gautam, Serhii Hamotskyi, Christian Hänig} 

\address{
         Anhalt University of Applied Sciences \\
         \{akash-kumar.gautam, serhii.hamotskyi, christian.haenig\}@hs-anhalt.de\\}

\abstract{
This paper describes team HSA\textunderscore CORAL's submission to the FinCausal 2026 shared task on extracting cause–effect relations from financial narratives via extractive question answering in English and Spanish. We compare three modeling families: (i) encoder-only token tagging with multilingual BERT, (ii) encoder–decoder generation with multilingual BART, and (iii) decoder-only LLMs (Llama 3.1 and GPT variants) using prompt refinement, few-shot demonstrations, and supervised fine-tuning. Across settings, prompting and few-shot examples yield competitive performance, while supervised fine-tuning provides the largest gains. Our best system, GPT-4.1 Mini fine-tuned on combined English and Spanish training data, achieves a tied highest score on the English subtask (score $4.8140$) and ranks third on Spanish (score $4.7753$) under the shared task's LLM-as-a-judge metric. Overall, the results highlight the value of task-specific adaptation and multilingual fine-tuning for cross-lingual transfer in financial causality QA. \\
\newline \Keywords{Large Language Models, Financial NLP,  Financial Question Answering.} }

\begin{document}

\maketitleabstract

\section{Introduction}
Understanding cause–effect relationships in financial texts is essential for informed decision-making. They reveal drivers of stock prices, economic changes, market behavior, and regulatory decisions across different countries. For analysts and investors, identifying causal connections provides valuable insights into financial risks, potential investments, and strategic planning \cite{gopalakrishnan2023text}. \\

The FinCausal shared task has steadily advanced how we detect causality in finance. Earlier editions focused on identifying causal phrases directly in text. Later versions introduced more complex challenges, such as recognizing implicit causality and multi-step reasoning \cite{mariko2022financial, mariko2020financial, zavitsanos2023financial}. The most recent task required generative models to answer open-ended questions about causes and effects, using Exact Match (EM) and Semantic Answer Similarity (SAS) \cite{risch2021semantic} as evaluation metrics \cite{moreno2025financial}. \\

The 2026 edition introduces several new elements \cite{moreno-sandoval-etal-2026-financial}. The dataset includes fragments with new complex causal relationships, rephrased questions that demand more sophisticated reasoning, and randomly divided train-test splits. A novel challenge is the use of an LLM-as-a-judge for evaluation, which rates responses on a 1--5 adequacy scale. The task pushes models beyond simple text matching toward understanding both explicit and implicit causal relationships in detailed financial narratives. \\

We explored three approaches to extractive question-answering: (i) token classification with encoder-only models like BERT, (ii) sequence-to-sequence generation with encoder-decoder models like BART, and (iii) few-shot prompting with decoder-only large language models. \\

Across our experiments, we find that fine-tuned generative models consistently outperform encoder-only and encoder-decoder architectures on the extractive QA task. Notably, GPT-4.1 Mini fine-tuned on a multilingual corpus achieves the best overall performance, securing the top position on the English subtask and third place on Spanish. Adding 20 few-shot examples proves critical, enabling even a smaller model to surpass its larger counterparts in zero-shot settings. These results demonstrate that multilingual fine-tuning and few-shot learning together provide a reliable approach for causal relation extraction in financial documents.

\section{Related Work}

\paragraph{Causal Information Extraction}
Early approaches to causal relationship detection in financial texts relied heavily on rule-based systems and traditional machine learning methods such as Support Vector Machines (SVMs) and decision trees \cite{ghosh2022lipi, baranes2019earning}. While these models demonstrated some success in identifying patterns in financial reports and news articles, they required extensive feature engineering and often struggled to capture the complexity and temporal dynamics of causal relations in domain-specific documents.

The introduction of BERT \cite{devlin2019bert} and its multilingual variants marked a significant shift in the field \cite{yang2019end,wan2022financial}. These pre-trained language models enabled more nuanced understanding of contextual information with minimal feature extraction, substantially improving performance on tasks involving document-level comprehension \citep{zhang2022hierarchical}. \\

Subsequent work has focused on fine-tuning these models on domain-specific financial datasets, achieving state-of-the-art results in tasks such as sentiment analysis and event extraction \cite{mariko2020financial}. Fine-tuned pre-trained language models have consistently outperformed traditional machine learning approaches, particularly when working with large-scale financial reports \cite{jin2023prototypical,huang2023finbert,sarmah2023towards}. \citet{liu2023event} propose an implicit cause–effect interaction framework to improve event causality extraction.\\

\paragraph{Pretrained Language Models}
More recently, proprietary large language models such as GPT-4 have been explored for question-answering tasks in the financial domain \cite{zhang2023moqagpt, kalpakchi2023quasi}. These models have demonstrated strong few-shot learning capabilities, enabling them to generalize from limited examples without task-specific fine-tuning \cite{xiao2022few,guo2023close}. This line of work underscores the growing potential of generative models for specialized NLP tasks, including the extraction of causal relationships in finance \cite{nayak2022generative, kim2023can}. \\

\begin{table*}[htbp]
\centering
% Set column widths
\newcolumntype{L}[1]{>{\raggedright\arraybackslash}p{#1}}
\newcolumntype{C}[1]{>{\centering\arraybackslash}p{#1}}

% Adjust row height
\renewcommand{\arraystretch}{1.3}

\begin{tabular}{
    L{3.5cm}       % Model column
    C{2.3cm}       % Fine-tuned column
    L{4cm}         % Corpus column
    C{1.5cm}       % English score column
    C{1.5cm}       % Spanish score column
}
\toprule
\textbf{Model} & 
\textbf{Fine-tuned} & 
\textbf{Corpus} & 
\textbf{English} & 
\textbf{Spanish} \\
\midrule
\small{BERT Base Multilingual} & yes & Multilingual (en+es) & 3.9800 & 3.9810 \\
\small{BART Facebook Base}         & yes & Multilingual (en+es) & 4.1200 & 4.0300 \\
\small{Llama-3.1 8B} &yes &Multilingual (en+es) & 4.0200 &3.9100 \\
\small{GPT-3.5 turbo}               & no  & -                      & 4.7040 & 4.7060 \\
\small{GPT-4.1 mini}                & yes & Monolingual (en)       & 4.7560 & 4.7141 \\
\small{GPT-4.1 mini}               & yes & Monolingual (es)       & 4.7210 & 4.7674 \\
\small{\textbf{GPT-4.1 mini}}               & yes & Multilingual (en+es) & \bf{4.8140} & \bf{4.7753} \\
\small{GPT-5.2}                   & no  & -                      & 4.7600   & 4.7350   \\
\bottomrule
\end{tabular}
\caption{Evaluation results on blinded English and Spanish test sets, with scores provided by an external LLM judge. The \textit{Corpus} column indicates whether models were fine-tuned on monolingual or multilingual data. Bold entries denote the highest score in each language column. Decoder-only models were tested with varying numbers of few-shot examples; results, obtained with 20 examples, are reported here. Best results using our approach are presented in \textbf{bold}.}
\label{tab:model_performance_booktabs}
\end{table*}

\section{Methodology}

We compare three approaches to extractive QA for FinCausal 2026: (1) token classification using BERT-based models, (2) extractive QA using encoder-decoder models, and (3) generative models. These approaches are compared across a variety of pretrained large language models in few-shot settings using a multilingual dataset.

\subsection{Encoder-Based Extractive QA}

This approach utilizes text embedding models such as BERT for token classification, following a similar methodology to \citet{yoon2022sequence}. The process begins by tokenizing both the context and the question, which are then concatenated with a special \texttt{[SEP]} token. During training, each sample is annotated using an IO tagging scheme, where the answer span is mapped to its first occurrence within the passage. \\

We compute the cross-entropy loss between the predicted token classes and the ground truth labels derived from the training data. To refine the loss calculation, a loss mask is applied to restrict the computation exclusively to tokens originating from the passage. This masking strategy excludes tokens from the question and special symbols (e.g., \texttt{[SEP]}, padding). This focuses learning on the passage tokens.

\subsection{Encoder-Decoder Extractive QA}

Our second approach treats extractive QA as a sequence-to-sequence generation task using BART \cite{lewis2020bart}. The input consists of the question and context concatenated with a delimiter, and the model is trained to generate the answer token-by-token. \\

We fine-tune BART by minimizing the negative log-likelihood of the target answer sequence. During inference, beam search is employed to decode the most likely answer. This formulation is effective for extractive tasks as it leverages BART's pretrained language understanding while flexibly handling answer spans. We evaluate multilingual BART variants on both the English and Spanish subtasks.

\subsection{Decoder-Based Extractive QA}
The flexibility of large language models makes them well-suited for extractive QA tasks. However, it is important to ensure that models follow instructions and avoid hallucinations beyond the provided context \cite{xu2024hallucination}. To address this, we implement a multi-step strategy combining prompt optimization, few-shot selection, and fine-tuning. \\

First, we perform prompt optimization through several iterative refinements on a small subset of the training dataset. The final version of the prompt used is described in Appendix \ref{sec:appendix_prompt}. This process helps identify instruction formats and phrasings that consistently yield extractive, context-grounded answers. We incorporate few-shot examples within the prompt, selecting relevant QA demonstrations using cosine similarity. Specifically, given a test context and question, we retrieve the most similar QA pairs from the training set based on multilingual embedding similarity\footnote{https://huggingface.co/sentence-transformers/paraphrase-multilingual-MiniLM-L12-v2}, and include them as examples in the prompt to guide the model's output.\\

As a further enhancement, we fine-tune the model on up to 2,000 samples, experimenting with three configurations: training on English data only, Spanish data only, and a combined bilingual dataset. Fine-tuning reinforces task-specific behavior and significantly reduces hallucinations, leading to more reliable extractive answers. \\

This hybrid approach—combining prompt engineering, dynamic few-shot selection, and targeted fine-tuning—allows us to leverage the generative strength of LLMs while maintaining the extractive precision required for the task.

\begin{figure}[t]
\begin{center}
\includegraphics[width=0.49\textwidth]{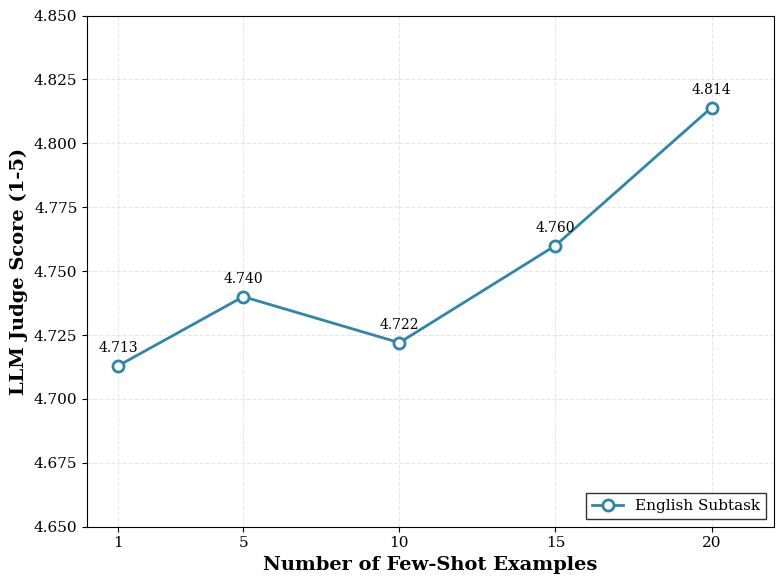}
\caption{External LLM-judge score (English subtask) as a function of the number of few-shot demonstrations in the prompt for the best-performing decoder-only configuration.}
\label{fig:few_shot_egs}
\end{center}
 \end{figure}

\section{Experiments}
\subsection{Dataset}
The dataset comprises financial narratives in English and Spanish designed for an extractive question answering task focused on causal relationships \citetlanguageresource{H7RKHH_2026}. Given a context and a question, the task requires identifying a text span that expresses a causal relation within the financial narrative~\cite{moreno2025financial}. 

Questions are formulated abstractly, targeting either the cause or the effect described in the text, with causes defined as agents or facts that can be extracted verbatim from the provided context. \\

For complex causal structures—such as causal chains or non-linear relationships—up to two questions are included per context to ensure comprehensive coverage. The English dataset is sourced from financial annual reports from 2017, drawn from the UCREL corpus and the English portion of the 2018 FinT-esp corpus, while the Spanish dataset is extracted from a corpus of Spanish financial annual reports spanning 2014 to 2018. The training set for both subtasks consists of 2,000 samples, with test sets of 500 samples for English and 503 samples for Spanish, respectively. 

\subsection{Model Selection}
BERT\footnote{https://huggingface.co/google-bert/bert-base-multilingual-cased} was used as an encoder model, multilingual BART\footnote{https://huggingface.co/facebook/bart-base} for encoder-decoder model and we evaluated Llama-3.1 and several GPT\footnote{https://developers.openai.com/api/docs/models} variants for decoder-only setups. For fine-tuning of Llama3.1 we used Low-Rank Adaptation (LoRA) to speed the process \cite{hu2022lora}. 

\subsection{Results}
\autoref{tab:model_performance_booktabs} reports the best-performing configuration for each model family. Scores correspond to the shared task’s external LLM-as-a-judge evaluation on the blind test set, rated on a 1--5 adequacy scale for both English and Spanish.

\paragraph{Encoder and Encoder-Decoder Setups.}
Extractive models such as BERT perform reasonably well at identifying exact spans corresponding to causal relationships, particularly when the answer is explicitly stated in the context. However, BART consistently outperforms BERT across both languages, benefiting from its sequence-to-sequence formulation which allows for more flexible and accurate span generation. Interestingly, the multilingual variant of BERT fine-tuned on combined English and Spanish data yields better performance on the blinded test set than its monolingual counterparts trained on individual languages. For brevity, we report only the best-performing configurations.

\paragraph{Few-Shot Learning with Decoder-Only Models}
In our decoder-only experiments, we evaluated Llama 3.1 and several variants of GPT. For fine-tuning Llama 3.1, we employed Low-Rank Adaptation (LoRA) to reduce computational overhead while maintaining task performance. Our findings indicate that while well-structured prompts with clear instructions are effective at reducing hallucinations, the inclusion of few-shot examples substantially improves the quality and precision of the generated answer spans. Notably, adding examples from the training set resulted in substantial improvements for GPT-4.1 Mini compared to zero-shot settings—even enabling it to outperform a much larger model (GPT-5.2\footnote{OpenAI currently supports supervised fine-tuning for GPT models up to GPT-4.1-2025-04-14.}) on both Spanish and English subtasks. \\

\autoref{fig:few_shot_egs} plots the LLM-as-a-judge scores against the increasing number of few-shot examples included in the prompt. The trend shows a clear improvement in output quality as more examples are added, with scores increasing correspondingly on the leaderboard. However, increasing the number of examples beyond a certain point did not yield further gains; in some cases, it even led to the generation of hallucinated content. \\

Given the context window constraints of each model, we systematically varied the number of few-shot examples. The best results across all decoder-only models were achieved with 20 few-shot examples, which we adopt as our final configuration.

\paragraph{Fine-Tuning}
Our experiments reveal a clear advantage of fine-tuned generative models over both encoder-only and encoder-decoder architectures. Multilingual fine-tuning consistently outperforms monolingual fine-tuning on both English and Spanish subtasks, demonstrating strong cross-lingual transfer for causal relation extraction in financial narratives. Among all configurations, GPT-4.1 Mini with multilingual fine-tuning achieved the best results across both languages. \\

Notably, this fine-tuned model outperforms a much larger model, GPT-5.2, in zero-shot settings. This suggests that the test dataset's contexts and questions contain lexical characteristics and causal relationships (for both Spanish and English) that can be reliably identified only when models have access to task-specific annotation guidelines through fine-tuning. The observation points to an interesting conclusion: fine-tuned pre-trained language models of moderate size may offer better performance for specialized tasks than much larger models used in inference-only mode, highlighting the importance of fine-tuning over parameter scaling for domain-specific applications. \\

\paragraph{Error Analysis and Limitations}
While our models achieved strong overall results, error analysis reveals persistent errors in handling nested causal structures and contexts containing multiple potential causes. In these challenging cases, models often selected the wrong causal pair, with errors more frequent in the Spanish subtask—suggesting that cross-lingual generalization remains challenging for complex causal structures.\\

The lack of detailed annotation guidelines for identifying relevant spans further compounds this issue, as prompt optimization alone cannot fully resolve such ambiguities. Looking ahead, alignment techniques such as Direct Preference Optimization \cite{rafailov2023direct} and Reinforcement Learning from Human Feedback \cite{bai2022training} offer promising ways for learning the implicit patterns underlying human annotations for structurally complex examples.

\section{Conclusion}
This work demonstrates that generative models outperform encoder and encoder-decoder architectures on causal question-answering tasks for financial narrative documents, provided that hallucinations are mitigated through careful prompt engineering and few-shot examples. Multilingual fine-tuning with a large number of training samples from both Spanish and English, significantly boosts performance, highlighting the effectiveness of cross-lingual transfer in this domain. Future work may explore LLM-based evaluation as a means to further enhance output quality, as well as more sophisticated strategies for handling nested and ambiguous causal structures.

\section*{Acknowledgments}
This work has been completed as part of project CORAL (Constrained Retrieval-Augmented Language Models), 
funded by the German Federal Ministry of Research, Technology, and Space (BMFTR) under grant number 16IS24077C.

%\renewcommand{\section}[1]{\oldsection{#1}}
% \renewcommand{\subsection}[1]{\oldsubsection{#1}}

%\section{Bibliographical References}\label{sec:reference},

\section{References}
\bibliographystyle{lrec2026-natbib}
\bibliography{ref}

\section{Language Resource References}
\label{lr:ref}
\bibliographystylelanguageresource{lrec2026-natbib}
\bibliographylanguageresource{languageresource}

\appendix

\section{Prompt}
\label{sec:appendix_prompt}
Figure \ref{fig:prompt_template} describes the prompt used by decoder only language models for generating the final response as part of the extractive question-asnwering task.

% Define a custom environment for the task box
\newmdenv[
  linecolor=gray!40,
  roundcorner=5pt,
  backgroundcolor=gray!5,
  skipabove=12pt,
  skipbelow=12pt,
  leftmargin=0pt,
  rightmargin=0pt,
  innertopmargin=10pt,
  innerbottommargin=10pt
]{taskbox}

% \begin{document}

\begin{figure*}[htbp]
\centering
\begin{minipage}{0.9\textwidth}
\begin{taskbox}
\small % Reduce font size slightly

\textbf{Task Description}

Given a financial context and a question, extract an exact answer from the context about cause or effect that addresses the question. The answer will be either the cause or the effect to a specific event mentioned in the context.

\vspace{0.5em}
\textbf{Instructions}

\begin{enumerate}
    \item \textbf{Read the context carefully:}
    \begin{itemize}
        \item Understand the events and relationship described in the context.
    \end{itemize}
    
    \item \textbf{Understand the question:}
    \begin{itemize}
        \item Determine if the question is asking for cause or effect.
        \item Identify the specific events or statement the question refers to.
    \end{itemize}
    
    \item \textbf{Extract the answer verbatim:}
    \begin{itemize}
        \item Locate the exact sentence or phrase in the context that answers the question.
        \item \textbf{Do not paraphrase, summarise, or add any external information. The answer must be copied word-for-word from the context.}
    \end{itemize}
    
    \item \textbf{Provide only the answer:}
    \begin{itemize}
        \item \textbf{Do not include any instructions, explanations or formatting.}
        \item \textbf{Output only the extracted answer and nothing else.}
    \end{itemize}
    
    \item \textbf{Answer should be in the language in which question is asked and the context is mentioned:}
    \begin{itemize}
        \item \textbf{Understand the context very well.}
    \end{itemize}
\end{enumerate}

\textbf{Examples}

\texttt{\{ \{ formatted\_examples \} \}}

\vspace{0.5em}
\textbf{Your Task}

\textit{Context:}

\texttt{\{ \{ context \} \}}

\textit{Question:}

\texttt{\{ \{ question \} \}}

\textit{Answer:}

[Provide only the exact answer from the context]

\vspace{0.5em}
\textbf{Remember}
\begin{itemize}
    \item \textbf{Output only the answer. Do not include any additional text. Do not include “Answer” in your output.}
    \item \textbf{The answer must exactly match a portion of the context.}
    \item \textbf{Do not add instructions, explanations, or any extra information.}
\end{itemize}
\end{taskbox}
\end{minipage}
\caption{Prompt used for extractive QA by decoder models used in the FinCausal 2026 shared task.}
\label{fig:prompt_template}
\end{figure*}

\end{document}